\pdfoutput=1

\documentclass[11pt]{article}
\usepackage{acl}

\usepackage{times}
\usepackage{latexsym}
\usepackage{multirow}
\usepackage{comment}

\usepackage[T1]{fontenc}
\usepackage[utf8]{inputenc}

\usepackage{microtype}

\usepackage{booktabs} 

\usepackage[cmex10]{amsmath}
\usepackage{amsfonts, amsthm, amssymb}
\usepackage{algorithm}
\usepackage{algpseudocode}
\usepackage{caption}
\usepackage{subcaption}

\usepackage{tikz}
\usetikzlibrary{arrows, decorations.pathreplacing, decorations.pathmorphing,backgrounds,positioning,fit,petri}

\theoremstyle{remark}

\def\A {\mathbf{A}}

\def\X {\mathbf{X}}

\def\c {\boldsymbol{c}}

\def\h {\boldsymbol{h}}

\def\o {\boldsymbol{o}}
\def\p {\boldsymbol{p}}

\def\s {\boldsymbol{s}}

\def\v {\boldsymbol{v}}
\def\w {\boldsymbol{w}}
\def\x {\boldsymbol{x}}
\def\y {\boldsymbol{y}}

\def\Tc {\mathcal{T}}

\def\Pb {\mathbb{P}}
\def\Eb {\mathbb{E}}

\usepackage[retainorgcmds]{IEEEtrantools}
\newcommand{\be}{\begin{IEEEeqnarray*}{rCl}}
\newcommand{\ee}{\end{IEEEeqnarray*}}

\newcommand{\ben}{\begin{IEEEeqnarray}{rCl}}
\newcommand{\een}{\end{IEEEeqnarray}}

\newcommand{\cb}[1]{\colorbox{gray!30}{#1}}

\title{A Flexible Multi-Task Model for BERT Serving}

\author{Tianwen Wei$^*$\quad\quad Jianwei Qi\thanks{\quad Equal contribution.}\quad\quad Shenghuan He\\
Xiaomi, XiaoAI Team \\
\texttt{\{weitianwen,qijianwei,heshenghuan\}@xiaomi.com}}

\begin{document}
\maketitle
\begin{abstract}
We present an efficient BERT-based multi-task (MT) framework that is particularly suitable for iterative and incremental development of the tasks.
The proposed framework is based on the idea of partial fine-tuning, i.e. only fine-tune some top layers of BERT while keep the other layers frozen. 
For each task,  we train independently a single-task (ST) model using partial fine-tuning. Then we compress the task-specific layers in each ST model using knowledge distillation.
Those compressed ST models are finally merged into one MT model so that the frozen layers of the former are shared across the tasks. 
 We exemplify our approach on eight GLUE tasks, demonstrating that it is able to achieve 99.6\% of the performance of the full fine-tuning method, while
 reducing up to two thirds of its overhead.  
 \end{abstract}

\section{Introduction}
In this work we explore the strategies of BERT \citep{BERT} serving for multiple tasks under the following two constraints:
1) Memory and computational resources are limited. On edge devices such as mobile phones, this is usually a hard constraint. On local GPU stations and Cloud-based servers, this constraint is not as hard but it is still desirable to reduce the computation overhead to cut the serving cost.
2) The tasks are expected to be modular and are subject to frequent updates. 
When one task is updated, the system should to be able to quickly adapt to the task modification such that the other tasks are not affected. 
This is a typical situation for applications (e.g. AI assistant) under iterative and incremental development.


In principle, there are two strategies of BERT serving: \emph{single-task serving} and \emph{multi-task serving}. 
In single-task serving, one independent single-task model is trained and deployed for each task. Typically, those models are obtained by fine-tuning a copy of the pre-trained BERT and are completely different from each other.
Single-task serving has the advantage of being flexible and modular as there is no dependency between the task models. 
The downside is its inefficiency in terms of both memory usage and computation, as neither parameters nor computation are shared or reused across the tasks. 
In multi-task serving, one single multi-task model is trained and deployed for all tasks. This model is typically trained with multi-task learning (MTL) \citep{caruana1997, RUDER2017}.
Compared to its single-task counterpart, multi-task serving is much more computationally efficient and incurs much less memory usage thanks to its sharing mechanism. However, it has the disadvantage in that any modification made to one task usually affect the other tasks. 

The main contribution of this work is the proposition of a framework for BERT serving that simultaneously achieves the flexibility of single-task serving and the efficiency of multi-task serving.
Our method is based on the idea of  partial fine-tuning, i.e. only fine-tuning some topmost layers of BERT  depending on the task and keeping the remaining bottom layers frozen.
The fine-tuned layers are task-specific, which can be updated on a per-task basis. The frozen layers at the bottom, which plays the role of a feature extractor, can be shared across the tasks.

\begin{table}[ht!]
\begin{center}
\resizebox{1.\columnwidth}{!}{%
\begin{tabular}{l|ccccccccc}
\toprule
$L$  & QNLI & RTE & QQP & MNLI & SST-2 & MRPC & CoLA & STS-B \\
\midrule 
1 & 85.9      & 60.3      & 86.1      & 77.1      & 91.6      & 77.2      & 38.7    & 84.8 \\
2 & 88.3      & 63.5      & 88.3      & 80.8      & 91.9      & 80.6      & 40.0    & 86.1 \\
3 & 89.9      & 65.3      & 89.0      & 82.5      & 91.2      & 84.6      & 45.3    & 87.3 \\
4 & 90.7      & 69.0      & 89.7      & 83.3      & 92.0      & 84.3      & 48.6    & \cb{88.2} \\
5 & \cb{91.0}      & {\bf \cb{71.5}}& {90.1} & \cb{84.0} & 92.2      & {\bf \cb{89.7}} & 51.3   & \cb{88.3} \\
6 & \cb{91.2} & \cb{71.1} & \cb{90.3} & \cb{84.2} & \cb{93.1} & 86.8      & 53.1    & 86.4 \\
7 & \cb{91.3}      & 70.0      & \cb{90.5} & \cb{83.9} & \cb{93.0} & 87.5      & 51.5    & \cb{88.6} \\
8 & \cb{91.5} & \cb{70.8} & \cb{90.6} & \cb{84.5} & \cb{92.8} & 88.0      & {\bf \cb{55.2}} & \cb{88.9} \\
9 & \cb{91.6} & \cb{70.8} & \cb{90.7} & \cb{84.0} & \cb{92.5} & 87.7      & \cb{54.7} & \cb{88.8} \\
10 & {\bf\cb{91.7}} & 69.7     & {\bf \cb{91.1}} & \cb{84.5} & \cb{93.0} & 87.3     & \cb{55.0} & \cb{88.7} \\
11 & {\bf\cb{91.7}} & 70.4     & {\bf \cb{91.1}} & \cb{84.5} & \cb{93.1} & 88.2      & \cb{54.7} & {\bf \cb{89.1}} \\
12 & \cb{91.6} & 69.7    & {\bf\cb{91.1}} & {\bf \cb{84.6}} & {\bf \cb{93.4}} & 88.2    & \cb{54.7} & \cb{88.8} \\
\bottomrule
\end{tabular}
}
\end{center}
\caption{Dev results on GLUE datasets obtained with partial fine-tuning. The parameter $L$ indicates the number of fine-tuned transformer layers. 
For each dataset and for each value of $L$, we always run the experiment 5 times with different initializations and report the maximum dev result obtained. 
 The best result in each column is highlighted in bold face. Shaded numbers indicate that they attain 99\% of the best result of the column. It can be seen that although fine-tuning more layers generally leads to better performance, the benefit of doing so suffers diminishing returns.  Perhaps surprisingly, for RTE, MRPC and CoLA it is the partial fine-tuning with roughly half of the layers frozen that gives the best results.
\label{partial_freezing}}
\end{table}

\section{Related Work}
The standard practice of using BERT is \emph{fine-tuning}, i.e. the entirety of the model parameters is adjusted on the training corpus of the downstream task, so that the model is adapted to that specific task \citep{BERT}. There is also an alternative \emph{feature-based} approach, used by ELMo \citep{ELMO2018}.
In the latter approach, the pre-trained model is regarded as a feature extractor with \emph{frozen parameters}. During the learning of a downstream task, one feeds a fixed or learnable combination of the model's intermediate representations as input to the task-specific module, and only the parameters of the latter will be updated.
It has been shown that the fine-tuning approach is generally superior to the {feature-based} approach for BERT in terms of task performance \citep{BERT, peters-etal-2019-tune}.

A natural middle ground between these two approaches is \emph{partial fine-tuning}, i.e. only fine-tuning some topmost layers of BERT while keeping the remaining bottom layers frozen. This approach has been studied in \citep{HOULSBY2019, merchant-etal-2020-happens}, where the authors observed that fine-tuning only the top layers can almost achieve the performance of full fine-tuning on several GLUE tasks.
The approach of partial fine-tuning essentially regards the bottom layers of BERT as a feature extractor. Freezing weights from bottom layers is a sensible idea as previous studies show
that the mid layer representations produced by BERT are most transferrable, whereas the top layers representations are more task-oriented \citep{wang2019, tenney2018, tenney2019, liu-etal-2019-linguistic, merchant-etal-2020-happens}. 
Notably, \citet{merchant-etal-2020-happens} showed that fine-tuning primarily affects weights from the top layers while weights from bottom layers do not alter much. 
 \citet{liu-etal-2019-linguistic} showed that it is possible to achieve state-of-the-art results on a number of probing tasks with linear models trained on frozen mid layer representations of BERT.


\begin{figure*}[ht!]
\centering
\begin{tikzpicture}
[cnode/.style={anchor=base, fill=#1, minimum size=6mm, rounded corners}, 
mnode/.style={anchor=base, fill=#1, text=brown, minimum size=6mm, rounded corners}, 
enode/.style={anchor=base, draw=#1, text=white, minimum size=6mm, rounded corners, dashed}, 
bnode/.style={anchor=base, fill=orange!30, minimum size=6mm}, 
>=stealth, scale=0.5]

\def\c{0.9}
\def\w{1.3}
\def\h{0.3}
\def\s{3}
\def\v{9}
\def\o{20.5}
\def\p{-5}


\foreach \y in {1,...,4}
{
\fill [gray!30] (- \w, \y*0.75+\h) rectangle (\w, \y*0.75-\h);
\node[] at (0, \y*0.75) {\tiny Layer {\tiny\tt \y}};
}

\foreach \y in {5,...,12}
{
\fill [blue!30] (- \w, \y*0.75+\h) rectangle (\w, \y*0.75-\h);
\node[] at (0, \y*0.75) {\tiny f.t. Layer {\tiny\tt \y}};
}

\node[rounded corners, thick, fill=blue!30] at (0, 14*0.75) {\small Task 1};
\draw[->, black, thick] (0, 12.5*0.75) -- (0, 13.4*0.75);

\foreach \y in {1,...,8}
{
\fill [gray!30] (\s- \w, \y*0.75+\h) rectangle (\s + \w, \y*0.75-\h);
\node[] at (\s, \y*0.75) {\tiny Layer {\tiny\tt \y}};
}

\foreach \y in {9, 10, 11,12}
{
\fill [red!30] (\s - \w, \y*0.75+\h) rectangle (\s+\w, \y*0.75-\h);
\node[] at (\s, \y*0.75) {\tiny f.t. Layer {\tiny\tt \y}};
}

\node[rounded corners, thick, fill=red!30] at (\s, 14*0.75) {\small Task 2};
\draw[->, black, thick] (\s, 12.5*0.75) -- (\s, 13.4*0.75);


\foreach \y in {1,...,4}
{
\fill [gray!30] (- \w + \v, \y*0.75+\h) rectangle (\w + \v, \y*0.75-\h);
\node[] at (\v, \y*0.75) {\tiny Layer {\tiny\tt \y}};
}

\foreach \y in {6, 9, 12}
{
\filldraw [blue!30] (- \w +\v, \y*0.75+\h - \h) rectangle (\w+\v, \y*0.75-\h - \h);
\node[] at (\v, \y*0.75 - \h) {\tiny Distill. Layer};
}

\node[rounded corners, thick, fill=blue!30] at (\v, 14*0.75) {\small Task 1};
\draw[->, black, thick] (\v, 12.5*0.75) -- (\v, 13.4*0.75);
\draw[thick, blue!50] (\v - 1.45, 12*0.75+\h) rectangle (\v + 1.45, 5*0.75 - \h + 0.2);

\foreach \y in {1,...,8}
{
\fill [gray!30] (\s- \w + \v, \y*0.75+\h) rectangle (\s + \w + \v, \y*0.75-\h);
\node[] at (\s + \v, \y*0.75) {\tiny Layer {\tiny\tt \y}};
}

\foreach \y in {10, 12}
{
\filldraw [red!30] (\s - \w +\v, \y*0.75+\h - \h) rectangle (\s+\w+\v , \y*0.75-\h - \h);
\node[] at (\s+\v, \y*0.75 - \h) {\tiny Distill. Layer};
}

\node[rounded corners, thick, fill=red!30] at (\s + \v, 14*0.75) {\small Task 2};
\draw[->, black, thick] (\s + \v, 12.5*0.75) -- (\s + \v, 13.4*0.75);
\draw[thick, red!50] (\s+\v - 1.45, 12*0.75+\h) rectangle (\s+\v + 1.45, 9*0.75 - \h + 0.2);


\foreach \y in {1,...,8}
{
\fill [gray!30] (- \w + \o, \y*0.75+\h) rectangle (\w +\o, \y*0.75-\h);
\node[] at (\o, \y*0.75) {\tiny Layer {\tiny\tt \y}};
}

\foreach \y in {10, 12}
{
\filldraw [red!30] (\s - \w +\o, \y*0.75+\h - \h) rectangle (\s+\w+\o , \y*0.75-\h - \h);
\node[] at (\s+\o, \y*0.75 - \h) {\tiny Distill. Layer};
}

\foreach \y in {6, 9, 12}
{
\filldraw [blue!30] (-\s - \w +\o, \y*0.75+\h - \h) rectangle (-\s+\w+\o, \y*0.75-\h - \h);
\node[] at (-\s+\o, \y*0.75 - \h) {\tiny Distill. Layer};
}

\node[rounded corners, thick, fill=blue!30] at (-\s+\o, 14*0.75) {\small Task 1};
\node[rounded corners, thick, fill=red!30] at (\s+\o, 14*0.75) {\small Task 2};

\draw[->, black, thick] (-1.4 + \o, 4*0.75) -- (-\s+\o,4*0.75) -- (-\s+\o, 5*0.75 - 0.2);	
\draw[->, black, thick] (1.4 + \o, 8*0.75) -- (\s+\o, 8*0.75) -- (\s+\o, 9*0.75 - 0.2);	

\draw[->, black, thick] (-\s+\o, 12.5*0.75) -- (-\s+\o, 13.4*0.75);
\draw[->, black, thick] (\s+\o, 12.5*0.75) -- (\s+\o, 13.4*0.75);

\draw[thick, blue!50] (-\s+\o - 1.5, 12*0.75+\h) rectangle (-\s+\o + 1.5, 5*0.75 - \h + 0.2);
\draw[thick, red!50] (\s+\o - 1.5, 12 *0.75 +\h) rectangle (\s+\o + 1.5, 9*0.75 - \h + 0.2);

\draw[->, black, double] (5, 7*0.75) -- (7, 7*0.75);
\node[] at (6, 7.5*0.75) {\small K.D.};

\draw[->, black, double] (14, 7*0.75) -- (15.5, 7*0.75);
\node[] at (14.7, 7.5*0.75) {\small Merge};

\node[] at (1.5, -0.5) {(a) Teacher models};
\node[] at (\v + 1.5, -0.5) {(b) Student models};
\node[] at (\o, -0.5) {(c) Final multi-task model};

\end{tikzpicture}
\caption{Pipeline of the proposed method. (a) For each task we train separately a task-specific model with partial fine-tuning, i.e. only the weights from some  topmost layers (blue and red blocks) of the pre-trained model are updated while the rest are kept frozen (gray blocks). 
(b) We perform knowledge distillation independently for each task on the task-specific layers of the teacher models.  
(c) The student models are merged into one MT model so that the frozen layers of the former can be shared.
 }
\label{model_architecture}
\end{figure*}
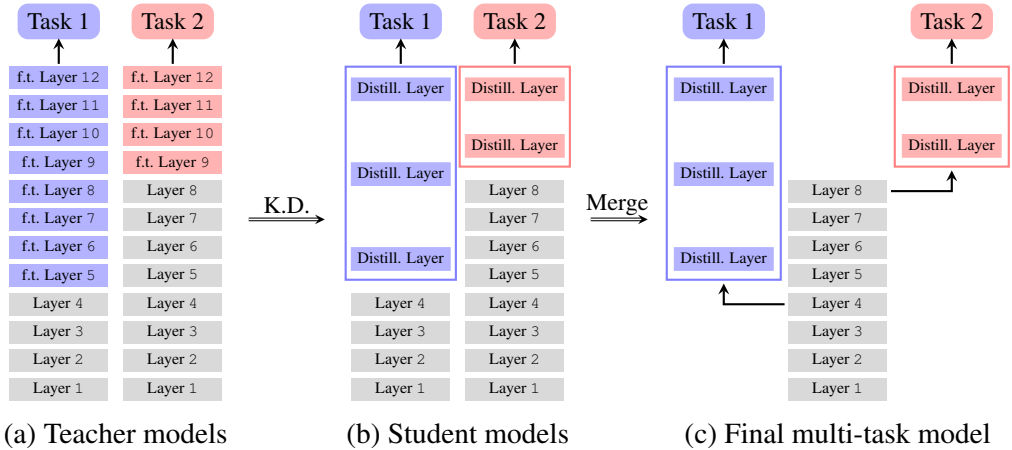

\section{Method}
In what follows, we denote by $\Tc$ the set of all target tasks.
  We always use the 12-layer uncased version of BERT as the pre-trained language model\footnote{The model checkpoint is downloaded from \url{https://storage.googleapis.com/bert_models/2018_10_18/uncased_L-12_H-768_A-12.zip}.}.
The proposed framework features a pipeline (Fig. \ref{model_architecture}) that consists of three steps: 1) Single task partial fine-tuning; 2) Single task knowledge distillation; 3) Model merging. We give details of these steps below.

\subsection{Single Task Partial Fine-Tuning \label{step1}}
In the first step, we partial fine-tune for each task an independent copy of BERT. The exact number of layers $L$ to fine-tune is a hyper-parameter and may vary across the tasks. We propose to experiment for each task with different value of $L$ within range  $N_{\min} \leqslant L \leqslant N_{\max}$,  and select the one that gives the best validation performance. 
The purpose of imposing the search range $[N_{\min}, N_{\max}]$ is to guarantee a minimum degree of parameter sharing. 
In the subsequent experiments on GLUE tasks (see Section \ref{results}), we set $N_{\min}=4$ and $N_{\max}=10$. 

This step produces a collection of single-task models as depicted in Fig. \ref{model_architecture}(a). We shall refer to them as single-task \emph{teacher models}, as they are to be knowledge distilled to further reduce the memory and computation overhead. 

\subsection{Single Task Knowledge Distillation \label{stkd}\label{step2}}
Since there is no interaction between the tasks, the process of knowledge distillation (KD) can be carried out separately for each task. In principle any of the existing KD methods for BERT \citep{MINILM, AGUILAR_KD, PATIENT_KD, TINY_BERT, BERT_THESEUS} suits our needs. In preliminary experiments we found out that as long as the student model is properly initialized, the vanilla knowledge distillation \citep{KD_HINTON} can be as performant as those more sophisticated methods. 

Assume that the teacher model for task $\tau\in\Tc$ contains $L^{(\tau)}$ fine-tuned layers at the top and $ 12 - L^{(\tau)}$ frozen layers at the bottom. Our goal is to compress the former into a smaller $l^{(\tau)}$-layer module. 
The proposed initialization scheme is very simple: we initialize the student model with the weights from the corresponding layers of the teacher. 
More precisely, let $N_s$ denote the number of layers (including both frozen and task-specific layers) in the student, where $N_s < 12$. We propose to initialize the student from the bottommost $N_s$ layers of the teacher.
Similar approach has also been used in \citep{distill_bert}, where the student is initialized by taking one layer out of two from the teacher. 
The value of $l^{(\tau)}$, i.e. the number of task-specific layers in the student model for task $\tau$, determines the final memory and computation overhead for that task. 


\subsection{Model Merging}
In the final step, we merge the single-task student models into one multi-task model (Fig. \ref{model_architecture}(c)) so that the parameters and computations carried out in the frozen layers can be shared.  To achieve this, it suffices to load weights from multiple model checkpoints into one computation graph.

\begin{table*}[ht!]
\begin{center}
\resizebox{2.\columnwidth}{!}{%
\begin{tabular}{l|cccccccc|c||c|c}
\toprule
                         & QNLI         & RTE          & QQP         & MNLI          & SST-2       & MRPC         & CoLA      & STS-B    & Avg. & Layers       & Overhead \\
\midrule 
Full fine-tuning          &  \cb{91.6}  & \cb{69.7}    & \cb{91.1}   & \cb{\bf 84.6} & \cb{93.4}   & \cb{88.2}    & \cb{54.7} & \cb{88.8}   & \cb{82.8} & $12\times8$  & 96 (100\%) \\
\midrule
DistillBERT$^{[b]}$      & {89.2}       & 59.9         & {88.5}      & {82.2}        & {91.3}      & \cb{87.5}    & {51.3}    & {86.9}   & 79.6 & $6\times8$ & 48 (50.0\%) \\
Vanilla-KD$^{[c]}$       & {88.0}       & 64.9         & {88.1}      & {80.1}        & {90.5}      & 86.2         & {45.1}    & {84.9}   & 78.5 & $6\times8$ & 48 (50.0\%) \\
PD-BERT$^{[d]}$          & {89.0}       & 66.7         & {89.1}      & {83.0}        & {91.1}      & 87.2         & {-}       & {-}      & -    & $6\times8$ & 48 (50.0\%) \\
BERT-PKD$^{[e]}$         & {88.4}       & 66.5         & {88.4}      & {81.3}        & {91.3}      & 85.7         & {45.5}    & {86.2}   & 79.2 & $6\times8$ & 48 (50.0\%) \\
BERT-of-Theseus$^{[f]}$  & {89.5}       & 68.2         & {89.6}      & {82.3}        & {91.5}      & \cb{89.0}    & {51.1}    & \cb{88.7}    & 81.2 & $6\times8$ & 48 (50.0\%) \\
TinyBERT$^{[g]}$         & \cb{90.5} & \cb{72.2}    & \cb{90.6} & {83.5} & {91.6} & \cb{88.4}    & {42.8} & {-}      & -    & $6\times8$ & 48 (50.0\%) \\
MiniLM$^{[h]}$           & {88.4} & 66.5    & {88.4} & {81.3} & {91.3} & 85.7    & {45.5} & {86.2}   & 79.2 & $6\times8$ & 48 (50.0\%) \\
\midrule
MT-DNN (full)$^{[j]}$    & \cb{91.1}    & \cb{\bf 80.9}& {87.6}      & \cb{84.4}     & \cb{\bf 93.5} & \cb{87.4}  & {51.3}    & {86.8}   & \cb{82.9} & $12\times1$ & {\bf 12} (12.5\%) \\
MT-DNN (LOO)$^{[k]}$     & {69.7}       & 60.6         & {66.5}      & {56.7}        & {79.2}        & 74.2       & {10.2}    & {72.9}   & - &  -  & -  \\
\midrule 
Ours (KD-1)           & 86.4            & 66.1         & \cb{91.0}   & {77.5}        & {90.7}        & 85.1       & {36.4}    & {88.3}   & 77.4 & $7+1\times$8 & 15 (15.6\%) \\
Ours (KD-2)           & 88.6            & 64.6         & \cb{\bf 91.3} & {81.7}      & \cb{92.7}     & 86.3       & {44.0}    & \cb{88.6}   & 79.7 & $7+2\times$8 & 23 (24.0\%)  \\
Ours (KD-3)           & 90.2            & 66.8         & \cb{91.2}   & {82.9}        & \cb{92.7}     & \cb{88.0}  & {50.0}    & \cb{\bf88.9}   & 81.3 & $7+3\times$8 & 31 (32.3\%) \\
\midrule
Ours (w/o KD)         & {\bf \cb{91.7}}  & \cb{71.5}    & \cb{91.1}  & \cb{84.5}     & \cb{93.1}     & \cb{\bf 89.7} & \cb{\bf 55.2} & \cb{\bf 88.9}   & \cb{\bf 83.2} & $7+60$ & 67 (69.8\%) \\
                      & ${\scriptstyle(2, 10)}$       & ${\scriptstyle(7, 5)}$     & ${\scriptstyle(2, 10)}$   & ${\scriptstyle(4, 8)}$   
                      & ${\scriptstyle(6, 6)}$    & ${\scriptstyle(7, 5)}$     & ${\scriptstyle(4, 8)}$    & ${\scriptstyle(4, 8)}$      &  &  &  \\
                      \midrule
Ours (mixed)          & 90.2            & \cb{71.5}    & \cb{91.0}   & {82.9}        & \cb{92.7}     & \cb{88.0}    & \cb{\bf 55.2} & \cb{88.3}   & \cb{82.5} & $7+26$ & 33 (34.3\%) \\
                      & ${\scriptstyle(2, 3)}$       & ${\scriptstyle(7, 5)}$     & ${\scriptstyle(2, 1)}$   & ${\scriptstyle(4, 3)}$   
                      & ${\scriptstyle(6, 2)}$    & ${\scriptstyle(7, 3)}$     & ${\scriptstyle(4, 8)}$    & ${\scriptstyle(4, 1)}$      &  &  &  \\
\bottomrule
\end{tabular}
}
\end{center}
\caption{A comparison of performance and overhead between our approach and various baselines (see \S \ref{baselines} for more details). 
The performance is evaluated on the dev set. To obtain the results labeled as ``Ours'', we always run the experiment 5 times with different initializations and report the maximum.
The best result in each column is highlighted in bold face. Shaded numbers indicate that they attain 99\% of the \emph{Full fine-tuning} baseline.
Results of $[b]$ are from \citep{distill_bert}; $[c]$-$[f]$ are from \citep{theseus}; $[g]$-$[h]$ are from \citep{MINILM}; $[j]$-$[k]$ are reproduced by us with the toolkit from \citep{MTDNN_2020}. 
Round bracket $(x, y)$ indicates that the underlying task model  before merging consists of $x$ frozen layers and $y$ task-specific layers (fine-tuned or knowledge-distilled).
In the ``Layers'' column, notation $7 + 2\times8$ implies that in the final multi-task model there are 7 shared frozen layers and 2 task-specific layers for each of the 8 task.
\label{comparison}}
\end{table*}

\section{Experiments \label{experiments}}
In this section, we compare the performance and efficiency of our model with various baselines on eight GLUE tasks \cite{wang2019glue}. More details on these tasks can be found in Appendix \ref{glue_datasets}. 
\subsection{Metrics}
The performance metrics for GLUE tasks is accuracy except for CoLA and STS-B.  We use Matthews correlation for CoLA, and Pearson correlation for STS-B. 

To measure the parameter and computational efficiency, we introduce the \emph{total number of transformer layers}
that are needed to perform inference for all eight tasks.
For the models studied in our experiments, the actual memory usage and the computational overhead are approximately linear with respect to
this number.
 It is named ``overhead'' in the header of Table \ref{comparison}.

\subsection{Baselines \label{baselines}}
The baseline models/methods can be divided into 4 categories:

\emph{ Single-task without KD.} There is only one method in this category, i.e. the standard practice of single task \emph{ full fine-tuning} that creates a separate model for each task.

\emph{ Single-task with KD.} The methods in this category create a separate model for each task, but a certain knowledge distillation method is applied to compress each task model into a 6-layer one. 
The KD methods include \citep{KD_HINTON, theseus, distill_bert, pd_bert, bert_pkd, TINY_BERT, MINILM}. 

\emph{ Multi-task learning.} This category includes two versions of MT-DNN \citep{MTDNN_2019, MTDNN_2020}, both of which produce one single multi-task model. 1) \emph{ MT-DNN (full)} is jointly trained for all eight tasks. It corresponds to the idea scenario where all tasks are known in advance.  2) \emph{ MT-DNN (LOO)}, where ``LOO'' stands for ``leave-one-out'',  corresponds to the scenario where one of the eight tasks is \emph{not} known in advance. 
The model is jointly pre-trained on the 7 available tasks. Then an output layer for the ``unknown'' task is trained with the pre-trained weights frozen. 

\emph{ Flexible multi-task.} Our models under various efficiency constraints. 
\emph{ Ours (w/o KD)} means that no knowledge distillation is applied to the task models.  
 The number of fine-tuned layers for each task is selected according to the criterion described in Section \ref{step1}.
\emph{ Ours (KD-$n$)} means that knowledge distillation is applied such that the student model for each task contains exactly $n$ task-specific  layers.  For \emph{ Ours (mixed)}, we determine the number of task-specific layers for each task based on the marginal benefit (in terms of task performance metric) of adding more layers to the task. More precisely, for each task we keep adding task-specific layers as long as the marginal benefit of doing so is no less than a  pre-determined threshold $c$.
In Table \ref{comparison}, we report the result for $c=1.0$. Results with other values of $c$ can be found in Appendix \ref{trade_off}.  

\subsection{Results\label{results}}
The results are summarized in Table \ref{comparison}. From the table it can be seen that the proposed method  \emph{ Ours (mixed)} outperforms all KD methods while being  more efficient. Compared to the single-task {full fine-tuning} baseline, our method reduces up to around two thirds of the total overhead while achieves 99.6\% of its performance. 


We observe that
MT-DNN (full) achieves the best average performance with the lowest overhead. 
However, its performance superiority primarily comes from one big boost on a single task (RTE) rather than consistent improvements on all tasks. 
In fact, we see that MT-DNN (full) suffers performance degradation on QQP and STS-B due to \emph{task interference}, a known problem for MTL \citep{caruana1997, bingel-2017, martinez-2017, wu2020}.
 From our perspective, the biggest disadvantage of MT-DNN
is that it assumes full knowledge of all target tasks in advance. From the results of MT-DNN (LOO), we observe that MT-DNN has difficulty in handling new tasks if the model is not allowed to be retrained.

\section{Discussions}
\subsection{Advantages}
One major advantage of the proposed architecture is its flexibility. 
First, different tasks may be fed with representations from different layers of BERT, which encapsulate different levels of linguistic information \citep{liu-etal-2019-linguistic}.
This flexibility is beneficial to both task performance and efficiency. 
For instance, on QQP we achieve an accuracy of 91.0, outperforming all KD baselines with \emph{merely one} task-specific layer (connected to the 2nd layer of the frozen backbone model). 
Second, our architecture explicitly allows for allocating uneven resources to different tasks.
 We have redistributed the resources among the tasks in \emph{ours (mixed)}, resulting in both greater performance and efficiency.
Third, our framework does not compromise the modular design of the system. The model can be straightforwardly updated on on a per-task basis.

\subsection{Limitations} \label{limitations}
The major limitation of our approach is that for each downstream task it requires approximately 10x more training time for the hyper-parameter search compared to the conventional approach. Although the cost is arguably manageable in practice, i.e. typically 2 or 3 days per task on a single Nvidia Tesla V100 GPU, the excessive computation load should not be overlooked.

Another limitation is that although the overall computation overhead is reduced, the \emph{serving latency} of our model deteriorates as the number of tasks grows, and may eventually be worse than that of the single task baseline. This is due to the fact that during inference one cannot get the output of any one task until the model has finished computing for \emph{all} tasks.
In this regard, our approach may not be appropriate for those applications that demand exceptionally low serving latency, e.g. below 10 ms. 
Nevertheless, we report in Appendix \ref{industrial_application}  an industrial use case where our multi-task model serves 21 tasks while achieving a 
latency as low as 32 ms (99th percentile).

\subsection{Comparison with Adaptor-Based Approaches}
The adaptor-based approaches \cite{HOULSBY2019, adapterhub} belong to another category of fine-tuning approaches that are also parameter-efficient.
 Basically, the adaptor-based approaches introduce one trainable task-specific ``adaptor'' module for each downstream task. This module is generally lightweight, containing only a few parameters and is inserted between (or within) layers of the backbone model (e.g. BERT). However, even though the parameters of the backbone model can be shared across the tasks, the computation for inference cannot due to the fact that the internal data flow in each task model is modified by the task-specific adaptor. Therefore, the adaptor-based approaches are not computationally efficient and one needs to perform a separate full forward pass for each task. Since both parameter and computation efficiency are what we aim to achieve, the adaptor-based approaches are not comparable to our method.

\section{Conclusion}
We have presented our framework that is designed to provide efficient and flexible BERT-based multi-task serving.
We have demonstrated on eight GLUE datasets that the proposed method achieves both strong performance and efficiency.
We release our code\footnote{\url{https://github.com/DandyQi/CentraBert}} and hope that it can facilitate BERT serving in cost-sensitive applications.

\bibliographystyle{acl_natbib}
\bibliography{nlp,myacl}

\begin{thebibliography}{39}
\expandafter\ifx\csname natexlab\endcsname\relax\def\natexlab#1{#1}\fi

\bibitem[{Aguilar et~al.(2020)Aguilar, Ling, Zhang, Yao, Fan, and
  Guo}]{AGUILAR_KD}
Gustavo Aguilar, Yuan Ling, Yu~Zhang, Benjamin Yao, Xing Fan, and Chenlei Guo.
  2020.
\newblock \href {https://aaai.org/ojs/index.php/AAAI/article/view/6229}
  {Knowledge distillation from internal representations}.
\newblock In \emph{The Thirty-Fourth {AAAI} Conference on Artificial
  Intelligence, {AAAI} 2020, The Thirty-Second Innovative Applications of
  Artificial Intelligence Conference, {IAAI} 2020, The Tenth {AAAI} Symposium
  on Educational Advances in Artificial Intelligence, {EAAI} 2020, New York,
  NY, USA, February 7-12, 2020}, pages 7350--7357. {AAAI} Press.

\bibitem[{Alonso and Plank(2017)}]{martinez-2017}
H{\'e}ctor Alonso and Barbara Plank. 2017.
\newblock \href {https://www.aclweb.org/anthology/E17-1005} {When is multitask
  learning effective? semantic sequence prediction under varying data
  conditions}.
\newblock In \emph{Proceedings of the 15th Conference of the {E}uropean Chapter
  of the Association for Computational Linguistics: Volume 1, Long Papers},
  pages 44--53, Valencia, Spain. Association for Computational Linguistics.

\bibitem[{Bentivogli et~al.(2009)Bentivogli, Clark, Dagan, and
  Giampiccolo}]{RTE}
Luisa Bentivogli, Peter Clark, Ido Dagan, and Danilo Giampiccolo. 2009.
\newblock The fifth pascal recognizing textual entailment challenge.
\newblock In \emph{TAC}.

\bibitem[{Bingel and S{o}gaard(2017)}]{bingel-2017}
Joachim Bingel and Anders S{o}gaard. 2017.
\newblock \href {https://www.aclweb.org/anthology/E17-2026} {Identifying
  beneficial task relations for multi-task learning in deep neural networks}.
\newblock In \emph{Proceedings of the 15th Conference of the {E}uropean Chapter
  of the Association for Computational Linguistics: Volume 2, Short Papers},
  pages 164--169, Valencia, Spain. Association for Computational Linguistics.

\bibitem[{Caruana(1997)}]{caruana1997}
Rich Caruana. 1997.
\newblock \href {https://doi.org/10.1023/A:1007379606734} {Multitask
  {Learning}}.
\newblock \emph{Machine Learning}, 28(1):41--75.
\newblock 00000.

\bibitem[{Cer et~al.(2017)Cer, Diab, Agirre, Lopez-Gazpio, and Specia}]{STSB}
Daniel Cer, Mona Diab, Eneko Agirre, I{\~n}igo Lopez-Gazpio, and Lucia Specia.
  2017.
\newblock \href {https://doi.org/10.18653/v1/S17-2001} {{S}em{E}val-2017 task
  1: Semantic textual similarity multilingual and crosslingual focused
  evaluation}.
\newblock In \emph{Proceedings of the 11th International Workshop on Semantic
  Evaluation ({S}em{E}val-2017)}, pages 1--14, Vancouver, Canada. Association
  for Computational Linguistics.

\bibitem[{Chen et~al.(2018)Chen, Zhang, Zhang, and ZHao}]{QQP}
Z.~Chen, H.~Zhang, X.~Zhang, and L.~ZHao. 2018.
\newblock Quora question pairs.

\bibitem[{Devlin et~al.(2019)Devlin, Chang, Lee, and Toutanova}]{BERT}
Jacob Devlin, Ming-Wei Chang, Kenton Lee, and Kristina Toutanova. 2019.
\newblock \href {https://doi.org/10.18653/v1/N19-1423} {{BERT}: Pre-training of
  deep bidirectional transformers for language understanding}.
\newblock In \emph{Proceedings of the 2019 Conference of the North {A}merican
  Chapter of the Association for Computational Linguistics: Human Language
  Technologies, Volume 1 (Long and Short Papers)}, pages 4171--4186,
  Minneapolis, Minnesota. Association for Computational Linguistics.

\bibitem[{Dolan and Brockett(2005)}]{MRPC}
William~B. Dolan and Chris Brockett. 2005.
\newblock \href {https://aclanthology.org/I05-5002} {Automatically constructing
  a corpus of sentential paraphrases}.
\newblock In \emph{Proceedings of the Third International Workshop on
  Paraphrasing ({IWP}2005)}.

\bibitem[{Hinton et~al.(2015)Hinton, Vinyals, and Dean}]{KD_HINTON}
Geoffrey Hinton, Oriol Vinyals, and Jeffrey Dean. 2015.
\newblock \href {http://arxiv.org/abs/1503.02531} {Distilling the knowledge in
  a neural network}.
\newblock In \emph{NIPS Deep Learning and Representation Learning Workshop}.

\bibitem[{Houlsby et~al.(2019)Houlsby, Giurgiu, Jastrzebski, Morrone,
  De~Laroussilhe, Gesmundo, Attariyan, and Gelly}]{HOULSBY2019}
Neil Houlsby, Andrei Giurgiu, Stanislaw Jastrzebski, Bruna Morrone, Quentin
  De~Laroussilhe, Andrea Gesmundo, Mona Attariyan, and Sylvain Gelly. 2019.
\newblock \href {http://proceedings.mlr.press/v97/houlsby19a.html}
  {Parameter-efficient transfer learning for {NLP}}.
\newblock In \emph{Proceedings of the 36th International Conference on Machine
  Learning}, volume~97 of \emph{Proceedings of Machine Learning Research},
  pages 2790--2799. PMLR.

\bibitem[{Jiao et~al.(2020)Jiao, Yin, Shang, Jiang, Chen, Li, Wang, and
  Liu}]{TINY_BERT}
Xiaoqi Jiao, Yichun Yin, Lifeng Shang, Xin Jiang, Xiao Chen, Linlin Li, Fang
  Wang, and Qun Liu. 2020.
\newblock \href {https://doi.org/10.18653/v1/2020.findings-emnlp.372}
  {{T}iny{BERT}: Distilling {BERT} for natural language understanding}.
\newblock In \emph{Findings of the Association for Computational Linguistics:
  EMNLP 2020}, pages 4163--4174, Online. Association for Computational
  Linguistics.

\bibitem[{Levesque et~al.(2012)Levesque, Davis, and Morgenstern}]{WNLI}
Hector Levesque, Ernest Davis, and Leora Morgenstern. 2012.
\newblock The winograd schema challenge.
\newblock In \emph{Thirteenth international conference on the principles of
  knowledge representation and reasoning}.

\bibitem[{Liu et~al.(2019{\natexlab{a}})Liu, Gardner, Belinkov, Peters, and
  Smith}]{liu-etal-2019-linguistic}
Nelson~F. Liu, Matt Gardner, Yonatan Belinkov, Matthew~E. Peters, and Noah~A.
  Smith. 2019{\natexlab{a}}.
\newblock \href {https://doi.org/10.18653/v1/N19-1112} {Linguistic knowledge
  and transferability of contextual representations}.
\newblock In \emph{Proceedings of the 2019 Conference of the North {A}merican
  Chapter of the Association for Computational Linguistics: Human Language
  Technologies, Volume 1 (Long and Short Papers)}, pages 1073--1094,
  Minneapolis, Minnesota. Association for Computational Linguistics.

\bibitem[{Liu et~al.(2019{\natexlab{b}})Liu, He, Chen, and Gao}]{MTDNN_2019}
Xiaodong Liu, Pengcheng He, Weizhu Chen, and Jianfeng Gao. 2019{\natexlab{b}}.
\newblock \href {https://doi.org/10.18653/v1/P19-1441} {Multi-task deep neural
  networks for natural language understanding}.
\newblock In \emph{Proceedings of the 57th Annual Meeting of the Association
  for Computational Linguistics}, pages 4487--4496, Florence, Italy.
  Association for Computational Linguistics.

\bibitem[{Liu et~al.(2020)Liu, Wang, Ji, Cheng, Zhu, Awa, He, Chen, Poon, Cao,
  and Gao}]{MTDNN_2020}
Xiaodong Liu, Yu~Wang, Jianshu Ji, Hao Cheng, Xueyun Zhu, Emmanuel Awa,
  Pengcheng He, Weizhu Chen, Hoifung Poon, Guihong Cao, and Jianfeng Gao. 2020.
\newblock \href {https://doi.org/10.18653/v1/2020.acl-demos.16} {The
  {M}icrosoft toolkit of multi-task deep neural networks for natural language
  understanding}.
\newblock In \emph{Proceedings of the 58th Annual Meeting of the Association
  for Computational Linguistics: System Demonstrations}, pages 118--126,
  Online. Association for Computational Linguistics.

\bibitem[{Merchant et~al.(2020)Merchant, Rahimtoroghi, Pavlick, and
  Tenney}]{merchant-etal-2020-happens}
Amil Merchant, Elahe Rahimtoroghi, Ellie Pavlick, and Ian Tenney. 2020.
\newblock \href {https://doi.org/10.18653/v1/2020.blackboxnlp-1.4} {What
  happens to {BERT} embeddings during fine-tuning?}
\newblock In \emph{Proceedings of the Third BlackboxNLP Workshop on Analyzing
  and Interpreting Neural Networks for NLP}, pages 33--44, Online. Association
  for Computational Linguistics.

\bibitem[{Peters et~al.(2018)Peters, Neumann, Iyyer, Gardner, Clark, Lee, and
  Zettlemoyer}]{ELMO2018}
Matthew Peters, Mark Neumann, Mohit Iyyer, Matt Gardner, Christopher Clark,
  Kenton Lee, and Luke Zettlemoyer. 2018.
\newblock \href {https://doi.org/10.18653/v1/N18-1202} {Deep contextualized
  word representations}.
\newblock In \emph{Proceedings of the 2018 Conference of the North {A}merican
  Chapter of the Association for Computational Linguistics: Human Language
  Technologies, Volume 1 (Long Papers)}, pages 2227--2237, New Orleans,
  Louisiana. Association for Computational Linguistics.

\bibitem[{Peters et~al.(2019)Peters, Ruder, and Smith}]{peters-etal-2019-tune}
Matthew~E. Peters, Sebastian Ruder, and Noah~A. Smith. 2019.
\newblock \href {https://doi.org/10.18653/v1/W19-4302} {To tune or not to tune?
  adapting pretrained representations to diverse tasks}.
\newblock In \emph{Proceedings of the 4th Workshop on Representation Learning
  for NLP (RepL4NLP-2019)}, pages 7--14, Florence, Italy. Association for
  Computational Linguistics.

\bibitem[{Pfeiffer et~al.(2020)Pfeiffer, R{\"u}ckl{\'e}, Poth, Kamath,
  Vuli{\'c}, Ruder, Cho, and Gurevych}]{adapterhub}
Jonas Pfeiffer, Andreas R{\"u}ckl{\'e}, Clifton Poth, Aishwarya Kamath, Ivan
  Vuli{\'c}, Sebastian Ruder, Kyunghyun Cho, and Iryna Gurevych. 2020.
\newblock \href {https://doi.org/10.18653/v1/2020.emnlp-demos.7}
  {{A}dapter{H}ub: A framework for adapting transformers}.
\newblock In \emph{Proceedings of the 2020 Conference on Empirical Methods in
  Natural Language Processing: System Demonstrations}, pages 46--54, Online.
  Association for Computational Linguistics.

\bibitem[{Radford et~al.(2018)Radford, Narasimhan, Salimans, and
  Sutskever}]{GPT1}
Alec Radford, Karthik Narasimhan, Tim Salimans, and Ilya Sutskever. 2018.
\newblock Improving language understanding by generative pre-training.

\bibitem[{Rajpurkar et~al.(2016)Rajpurkar, Zhang, Lopyrev, and Liang}]{SQUAD}
Pranav Rajpurkar, Jian Zhang, Konstantin Lopyrev, and Percy Liang. 2016.
\newblock \href {http://arxiv.org/abs/1606.05250} {Squad: 100,000+ questions
  for machine comprehension of text}.

\bibitem[{Ruder(2017)}]{RUDER2017}
Sebastian Ruder. 2017.
\newblock \href {http://arxiv.org/abs/1706.05098} {An overview of multi-task
  learning in deep neural networks}.
\newblock \emph{CoRR}, abs/1706.05098.

\bibitem[{Sanh et~al.(2019)Sanh, Debut, Chaumond, and Wolf}]{distill_bert}
Victor Sanh, Lysandre Debut, Julien Chaumond, and Thomas Wolf. 2019.
\newblock \href {http://arxiv.org/abs/1910.01108} {Distilbert, a distilled
  version of bert: smaller, faster, cheaper and lighter}.
\newblock \emph{CoRR}, abs/1910.01108.

\bibitem[{Socher et~al.(2013)Socher, Perelygin, Wu, Chuang, Manning, Ng, and
  Potts}]{SST2}
Richard Socher, Alex Perelygin, Jean Wu, Jason Chuang, Christopher~D. Manning,
  Andrew Ng, and Christopher Potts. 2013.
\newblock \href {https://aclanthology.org/D13-1170} {Recursive deep models for
  semantic compositionality over a sentiment treebank}.
\newblock In \emph{Proceedings of the 2013 Conference on Empirical Methods in
  Natural Language Processing}, pages 1631--1642, Seattle, Washington, USA.
  Association for Computational Linguistics.

\bibitem[{Sun et~al.(2019{\natexlab{a}})Sun, Cheng, Gan, and Liu}]{PATIENT_KD}
Siqi Sun, Yu~Cheng, Zhe Gan, and Jingjing Liu. 2019{\natexlab{a}}.
\newblock \href {https://doi.org/10.18653/v1/D19-1441} {Patient knowledge
  distillation for {BERT} model compression}.
\newblock In \emph{Proceedings of the 2019 Conference on Empirical Methods in
  Natural Language Processing and the 9th International Joint Conference on
  Natural Language Processing (EMNLP-IJCNLP)}, pages 4323--4332, Hong Kong,
  China. Association for Computational Linguistics.

\bibitem[{Sun et~al.(2019{\natexlab{b}})Sun, Cheng, Gan, and Liu}]{bert_pkd}
Siqi Sun, Yu~Cheng, Zhe Gan, and Jingjing Liu. 2019{\natexlab{b}}.
\newblock \href {https://doi.org/10.18653/v1/D19-1441} {Patient knowledge
  distillation for {BERT} model compression}.
\newblock In \emph{Proceedings of the 2019 Conference on Empirical Methods in
  Natural Language Processing and the 9th International Joint Conference on
  Natural Language Processing (EMNLP-IJCNLP)}, pages 4323--4332, Hong Kong,
  China. Association for Computational Linguistics.

\bibitem[{Tenney et~al.(2019{\natexlab{a}})Tenney, Das, and
  Pavlick}]{tenney2019}
Ian Tenney, Dipanjan Das, and Ellie Pavlick. 2019{\natexlab{a}}.
\newblock \href {https://doi.org/10.18653/v1/P19-1452} {{BERT} rediscovers the
  classical {NLP} pipeline}.
\newblock In \emph{Proceedings of the 57th Annual Meeting of the Association
  for Computational Linguistics}, pages 4593--4601, Florence, Italy.
  Association for Computational Linguistics.

\bibitem[{Tenney et~al.(2019{\natexlab{b}})Tenney, Xia, Chen, Wang, Poliak,
  McCoy, Kim, Durme, Bowman, Das, and Pavlick}]{tenney2018}
Ian Tenney, Patrick Xia, Berlin Chen, Alex Wang, Adam Poliak, R~Thomas McCoy,
  Najoung Kim, Benjamin~Van Durme, Sam Bowman, Dipanjan Das, and Ellie Pavlick.
  2019{\natexlab{b}}.
\newblock \href {https://openreview.net/forum?id=SJzSgnRcKX} {What do you learn
  from context? probing for sentence structure in contextualized word
  representations}.
\newblock In \emph{International Conference on Learning Representations}.

\bibitem[{Turc et~al.(2019)Turc, Chang, Lee, and Toutanova}]{pd_bert}
Iulia Turc, Ming-Wei Chang, Kenton Lee, and Kristina Toutanova. 2019.
\newblock \href {http://arxiv.org/abs/1908.08962} {Well-read students learn
  better: On the importance of pre-training compact models}.

\bibitem[{Wang et~al.(2019{\natexlab{a}})Wang, Hula, Xia, Pappagari, McCoy,
  Patel, Kim, Tenney, Huang, Yu, Jin, Chen, Van~Durme, Grave, Pavlick, and
  Bowman}]{wang2019}
Alex Wang, Jan Hula, Patrick Xia, Raghavendra Pappagari, R.~Thomas McCoy, Roma
  Patel, Najoung Kim, Ian Tenney, Yinghui Huang, Katherin Yu, Shuning Jin,
  Berlin Chen, Benjamin Van~Durme, Edouard Grave, Ellie Pavlick, and Samuel~R.
  Bowman. 2019{\natexlab{a}}.
\newblock \href {https://doi.org/10.18653/v1/P19-1439} {Can you tell me how to
  get past sesame street? sentence-level pretraining beyond language modeling}.
\newblock In \emph{Proceedings of the 57th Annual Meeting of the Association
  for Computational Linguistics}, pages 4465--4476, Florence, Italy.
  Association for Computational Linguistics.

\bibitem[{Wang et~al.(2018)Wang, Singh, Michael, Hill, Levy, and Bowman}]{GLUE}
Alex Wang, Amanpreet Singh, Julian Michael, Felix Hill, Omer Levy, and Samuel
  Bowman. 2018.
\newblock \href {https://doi.org/10.18653/v1/W18-5446} {{GLUE}: A multi-task
  benchmark and analysis platform for natural language understanding}.
\newblock In \emph{Proceedings of the 2018 {EMNLP} Workshop {B}lackbox{NLP}:
  Analyzing and Interpreting Neural Networks for {NLP}}, pages 353--355,
  Brussels, Belgium. Association for Computational Linguistics.

\bibitem[{Wang et~al.(2019{\natexlab{b}})Wang, Singh, Michael, Hill, Levy, and
  Bowman}]{wang2019glue}
Alex Wang, Amanpreet Singh, Julian Michael, Felix Hill, Omer Levy, and Samuel~R
  Bowman. 2019{\natexlab{b}}.
\newblock Glue: A multi-task benchmark and analysis platform for natural
  language understanding.
\newblock In \emph{7th International Conference on Learning Representations,
  ICLR 2019}.

\bibitem[{Wang et~al.(2020)Wang, Wei, Dong, Bao, Yang, and Zhou}]{MINILM}
Wenhui Wang, Furu Wei, Li~Dong, Hangbo Bao, Nan Yang, and Ming Zhou. 2020.
\newblock \href
  {https://proceedings.neurips.cc/paper/2020/hash/3f5ee243547dee91fbd053c1c4a845aa-Abstract.html}
  {Minilm: Deep self-attention distillation for task-agnostic compression of
  pre-trained transformers}.
\newblock In \emph{Advances in Neural Information Processing Systems 33: Annual
  Conference on Neural Information Processing Systems 2020, NeurIPS 2020,
  December 6-12, 2020, virtual}.

\bibitem[{Warstadt et~al.(2018)Warstadt, Singh, and Bowman}]{COLA}
A.~Warstadt, A.~Singh, and S.~R. Bowman. 2018.
\newblock Corpus of linguistic acceptability.

\bibitem[{Williams et~al.(2018)Williams, Nangia, and Bowman}]{MNLI}
Adina Williams, Nikita Nangia, and Samuel Bowman. 2018.
\newblock \href {https://doi.org/10.18653/v1/N18-1101} {A broad-coverage
  challenge corpus for sentence understanding through inference}.
\newblock In \emph{Proceedings of the 2018 Conference of the North {A}merican
  Chapter of the Association for Computational Linguistics: Human Language
  Technologies, Volume 1 (Long Papers)}, pages 1112--1122, New Orleans,
  Louisiana. Association for Computational Linguistics.

\bibitem[{Wu et~al.(2020)Wu, Zhang, and R{\'e}}]{wu2020}
Sen Wu, Hongyang~R Zhang, and Christopher R{\'e}. 2020.
\newblock Understanding and improving information transfer in multi-task
  learning.
\newblock \emph{arXiv preprint arXiv:2005.00944}.

\bibitem[{Xu et~al.(2020{\natexlab{a}})Xu, Zhou, Ge, Wei, and
  Zhou}]{BERT_THESEUS}
Canwen Xu, Wangchunshu Zhou, Tao Ge, Furu Wei, and Ming Zhou.
  2020{\natexlab{a}}.
\newblock \href {https://doi.org/10.18653/v1/2020.emnlp-main.633}
  {{BERT}-of-theseus: Compressing {BERT} by progressive module replacing}.
\newblock In \emph{Proceedings of the 2020 Conference on Empirical Methods in
  Natural Language Processing (EMNLP)}, pages 7859--7869, Online. Association
  for Computational Linguistics.

\bibitem[{Xu et~al.(2020{\natexlab{b}})Xu, Zhou, Ge, Wei, and Zhou}]{theseus}
Canwen Xu, Wangchunshu Zhou, Tao Ge, Furu Wei, and Ming Zhou.
  2020{\natexlab{b}}.
\newblock \href {https://doi.org/10.18653/v1/2020.emnlp-main.633}
  {{BERT}-of-theseus: Compressing {BERT} by progressive module replacing}.
\newblock In \emph{Proceedings of the 2020 Conference on Empirical Methods in
  Natural Language Processing (EMNLP)}, pages 7859--7869, Online. Association
  for Computational Linguistics.

\end{thebibliography}
\appendix
\pagebreak
\begin{center}
\textbf{\large Supplemental Materials}
\end{center}
\section{Details on the GLUE tasks \label{glue_datasets}}
The GLUE benchmark includes the following datasets:
\begin{itemize}
\item{\bf QNLI} (Question Natural Language Inference). The dataset is derived from \cite{SQUAD}. This is a binary classification task where an example is of the form (question, sentence) and the goal is to predict whether the sentence contains the correct answer to the question \cite{GLUE}.
\item{\bf RTE} (Recognizing Textual Entailment). A binary entailment task similar to MNLI but with much less training data \cite{RTE}.

\item{\bf QQP} (Quora Question Pairs) A binary classification task where the goal is to determine if two questions asked on Quora are semantically equivalent \cite{QQP}.
\item{\bf MNLI} (Multi-Genre Natural Language Inference). Given a pair of sentences, the goal is to predict whether the second sentence is an entailment, contradiction or neutral with respect to the first one \cite{MNLI}.
\item{\bf SST-2} (The Stanford Sentiment Treebank). A binary single-sentence classification task where the goal is to predict the sentiment (positive or negative) of the movie reviews \cite{SST2}.
\item{\bf MRPC} (Microsoft Research Paraphrase Corpus). A binary classification task where the goal is to predict whether two sentences are semantically equivalent \cite{MRPC}.
\item{\bf CoLA} (The Corpus of Linguistic Acceptability). A binary single-sentence classification task where the goal is to predict whether an English sentence is linguistically ``acceptable'' or not \cite{COLA}.
\item{\bf STS-B} (The Semantic Textual Similarity Benchmark). A regression task where the goal is to predict whether two sentences are similar in terms of semantic meaning as measured by a score from 1 to 5 \cite{STSB}.
\item{\bf WNLI} (Winograd NLI). The dataset is derived from \cite{WNLI}. We exclude this task in our experiments following the practice of \cite{BERT, GPT1}.
\end{itemize}
\begin{table}[ht!]
\begin{center}
\begin{tabular}{l|rr}
\toprule
Dataset & Train & Dev \\
\midrule
QNLI & 108k & 5.4k \\
RTE & 2.5k &  0.3k \\
QQP & 363k & 40k \\
MNLI & 392k & 9.8k \\
SST-2 & 67k & 0.8k \\
MRPC & 3.5k & 0.4k \\
CoLA & 8.5k & 1.0k \\
STS-B  & 5.7k & 1.5k \\ 
\bottomrule
\end{tabular}
\end{center}
\caption{Number of examples for training and development in GLUE datasets.}
\end{table}

\section{Hyper-parameters \label{hyper_parameters}}
The approach presented in this work introduces two new hyper-parameters for each task $\tau\in\Tc$, namely the number of fine-tuned layers $L^{(\tau)}$ for the teacher and the number of knowledge distilled layer $l^{(\tau)}$ for the student. If the resources permit, these two hyper-parameters should be tuned separately for each task. As introduced in Section \ref{step1}, we suggest to constrain $L$ within the range $4\leq L^{(\tau)} \leq 10$. As for $l^{(\tau)}$ which determines the eventual task-specific overhead, we impose $l^{(\tau)}\leq 3$. Since we always determines $L^{(\tau)}$ first, we do not need to experiment with every combination of $(L^{(\tau)}, l^{(\tau)})$. 
Combining these together, our approach requires approximately 10x (7 for $L$ and 3 for $l$) more training time compared to conventional full fine-tuning approach. 

The conventional hyper-parameters (e.g. learning rate, mini-batch size, etc) used in our experiments are summarized in Table \ref{hyperparameters}.
\begin{table}[ht!]
\begin{center}
\begin{tabular}{c|c}
\toprule
Hyper-parameter & value \\
\midrule
learning rate  & 2e-5  \\
batch size  & 32 \\
Epoch  & 3, 4, 5 \\
Optimizer & Adam \\
weight decay rate  & 0.01  \\
$\beta_1$  & 0.9 \\
$\beta_2$  & 0.999 \\
$\epsilon$ & 1e-6  \\
\bottomrule
\end{tabular}
\end{center}
\caption{Hyper-parameters used in our experiments. We mainly followed the practice of \cite{BERT}.   \label{hyperparameters}}
\end{table}

\section{Detailed Experiment Results\label{detailed_results}}
\begin{figure*}
     \centering
     \begin{subfigure}[b]{0.45\textwidth}
         \centering
         \includegraphics[width=\textwidth]{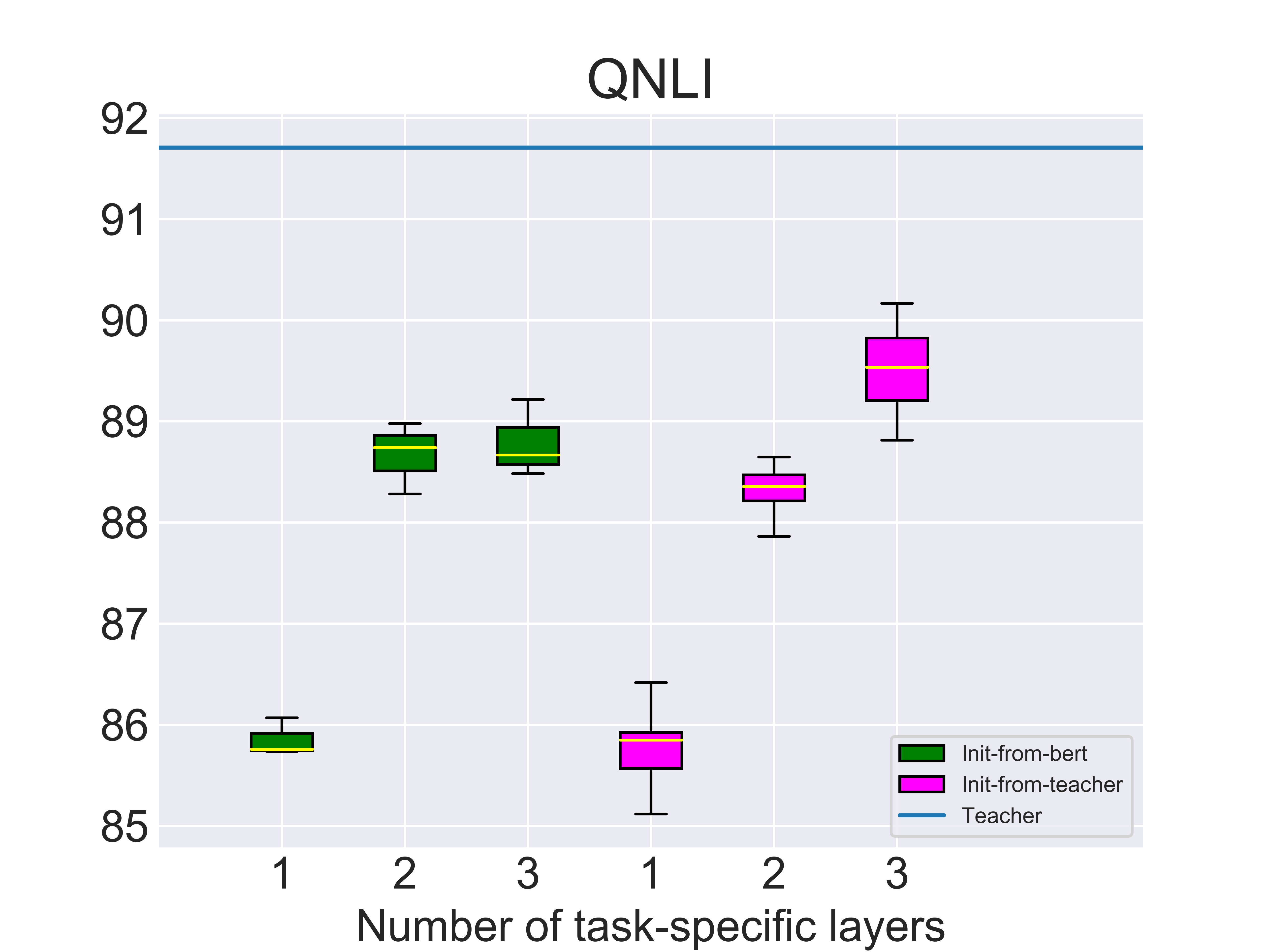}
     \end{subfigure}
     \begin{subfigure}[b]{0.45\textwidth}
         \centering
         \includegraphics[width=\textwidth]{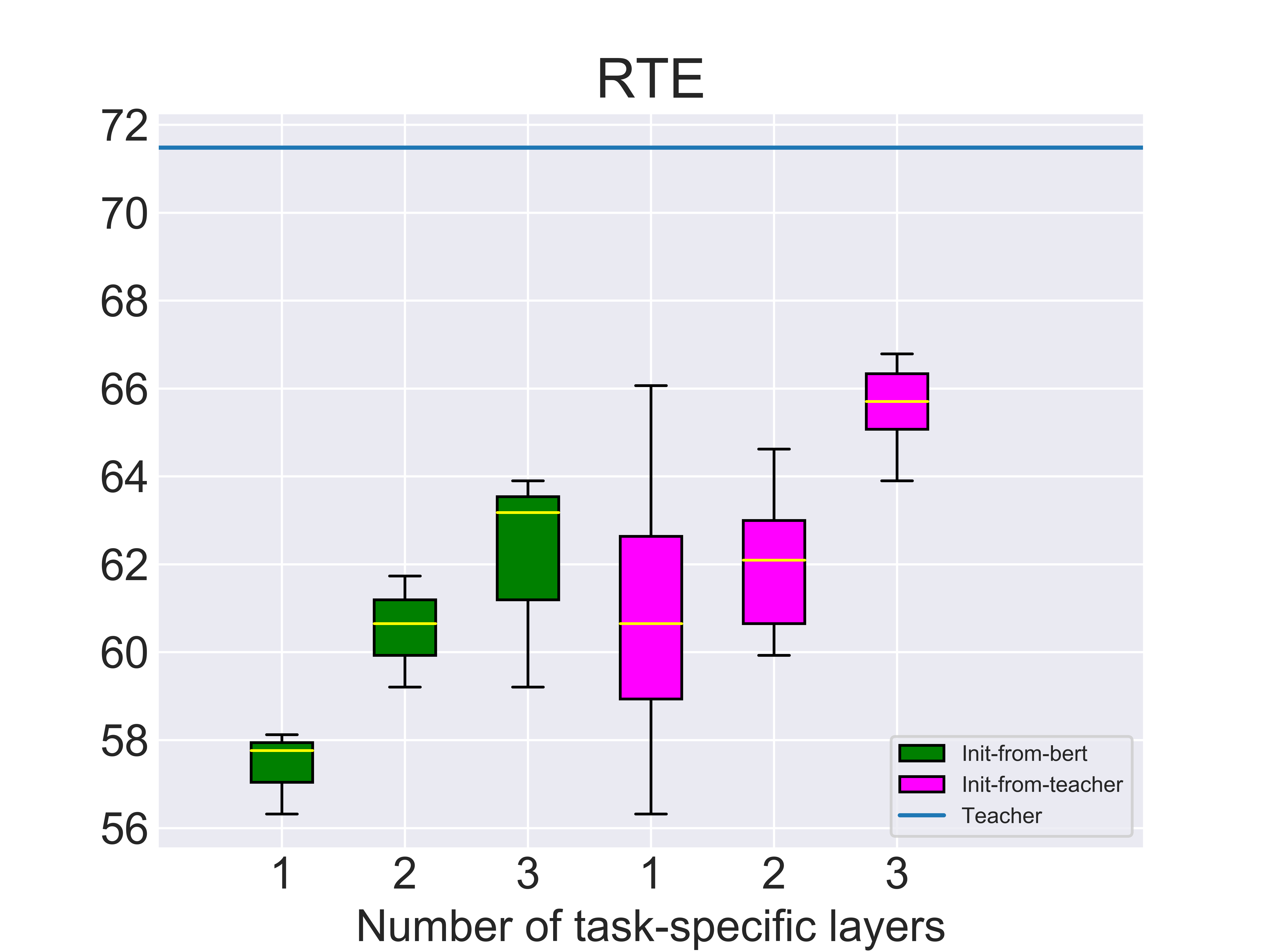}
     \end{subfigure}
     
     \begin{subfigure}[b]{0.45\textwidth}
         \centering
         \includegraphics[width=\textwidth]{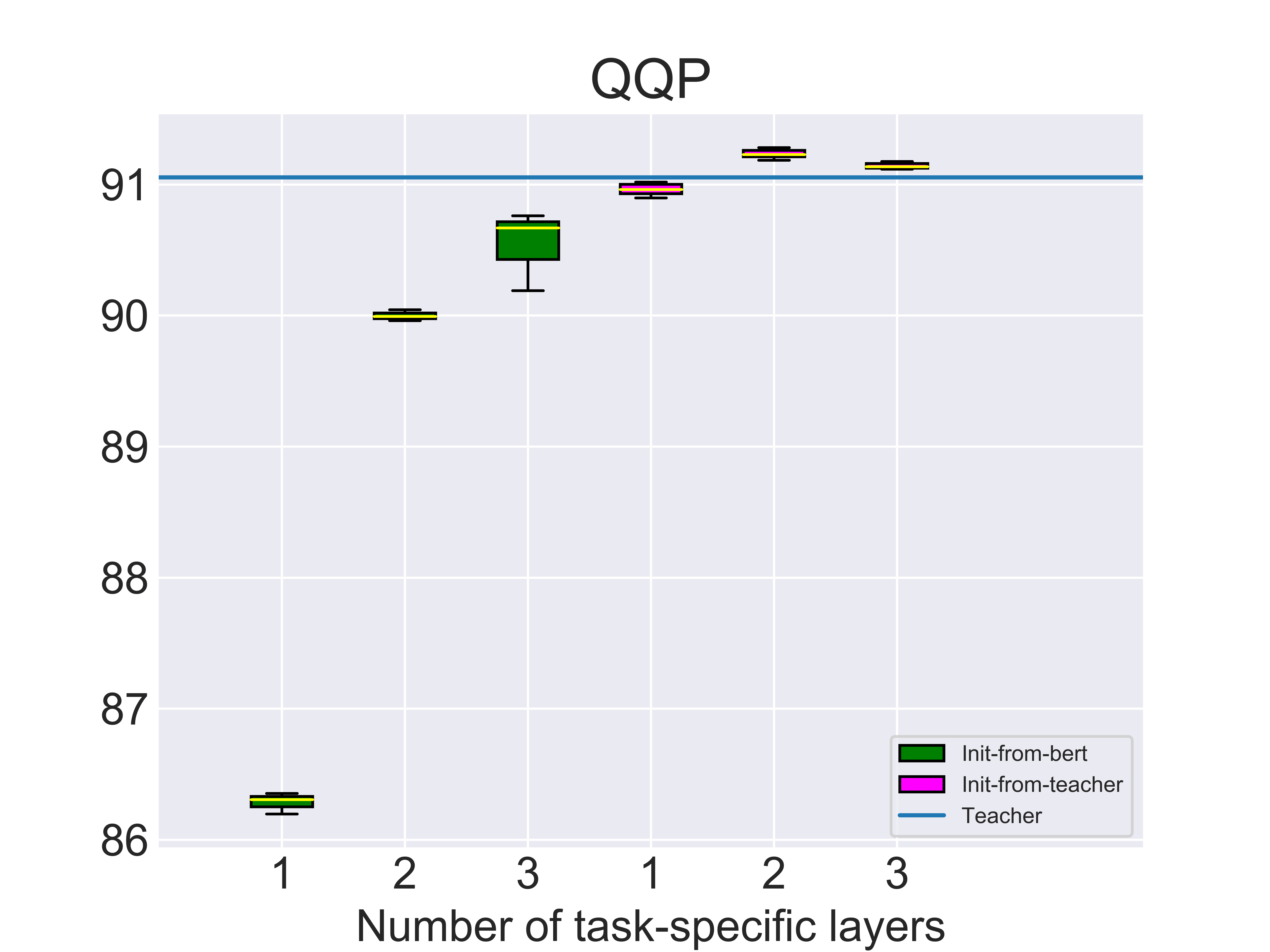}
     \end{subfigure}
     \begin{subfigure}[b]{0.45\textwidth}
         \centering
         \includegraphics[width=\textwidth]{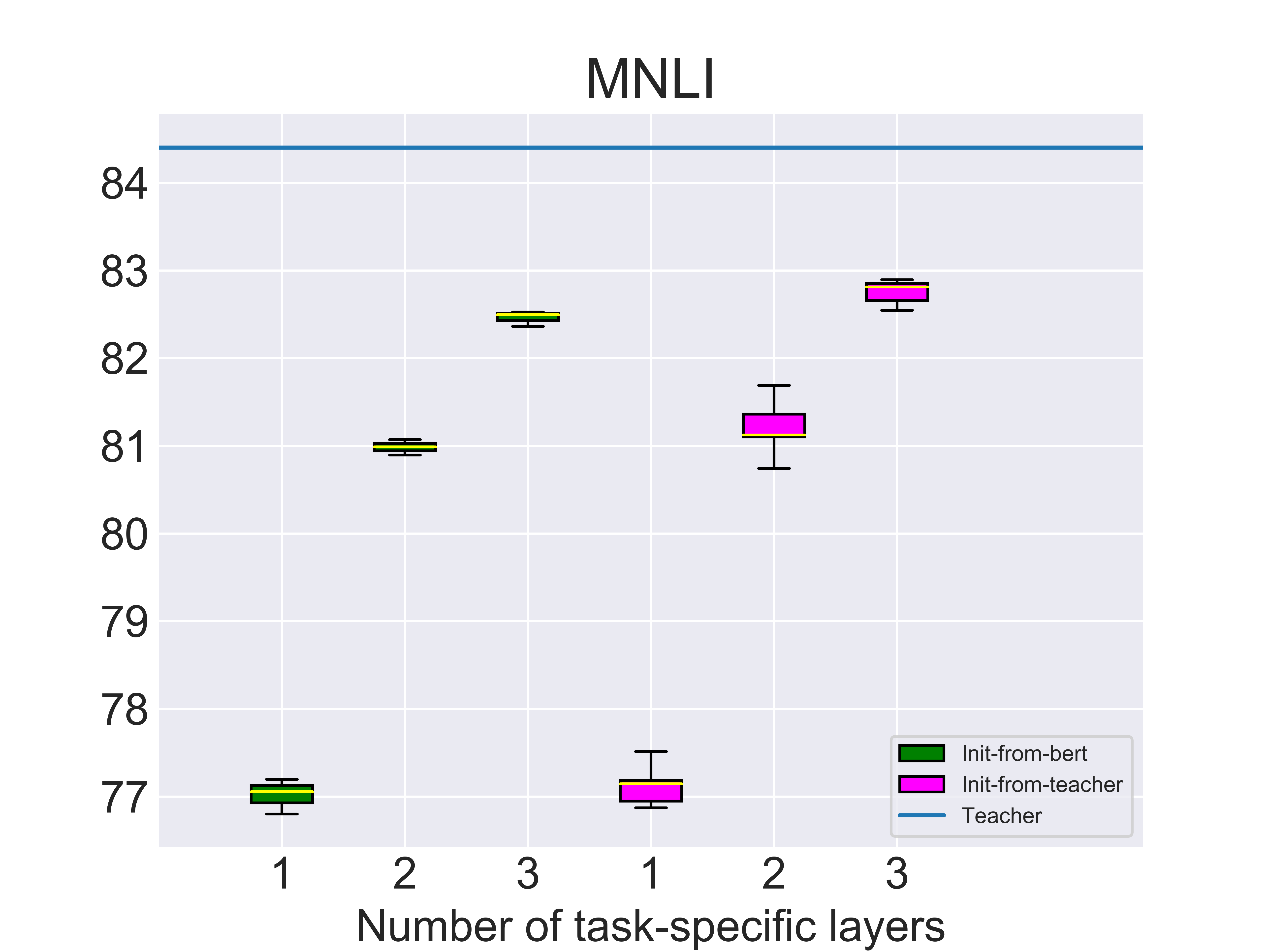}
     \end{subfigure}

\begin{subfigure}[b]{0.45\textwidth}
         \centering
         \includegraphics[width=\textwidth]{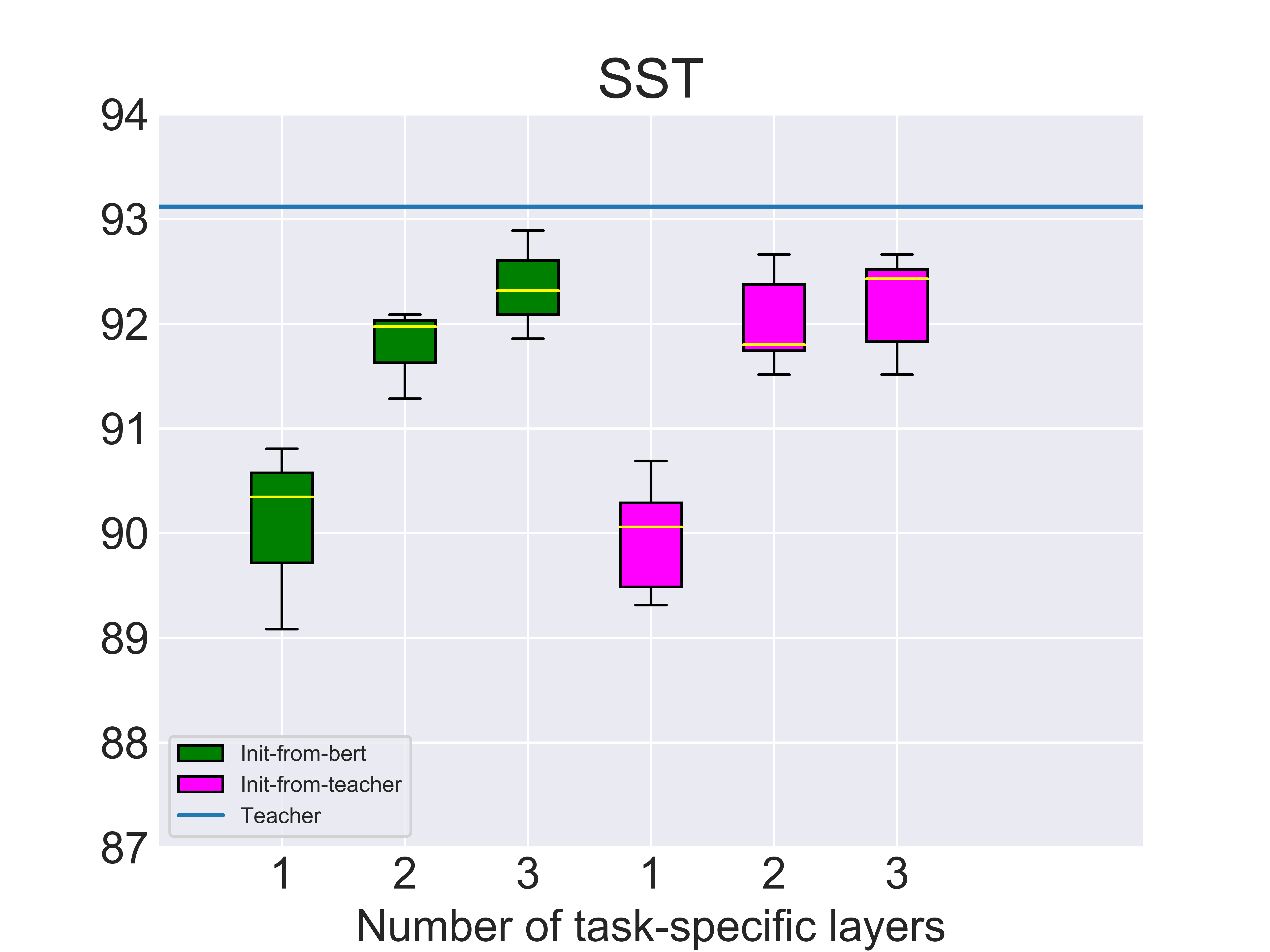}
     \end{subfigure}
     \begin{subfigure}[b]{0.45\textwidth}
         \centering
         \includegraphics[width=\textwidth]{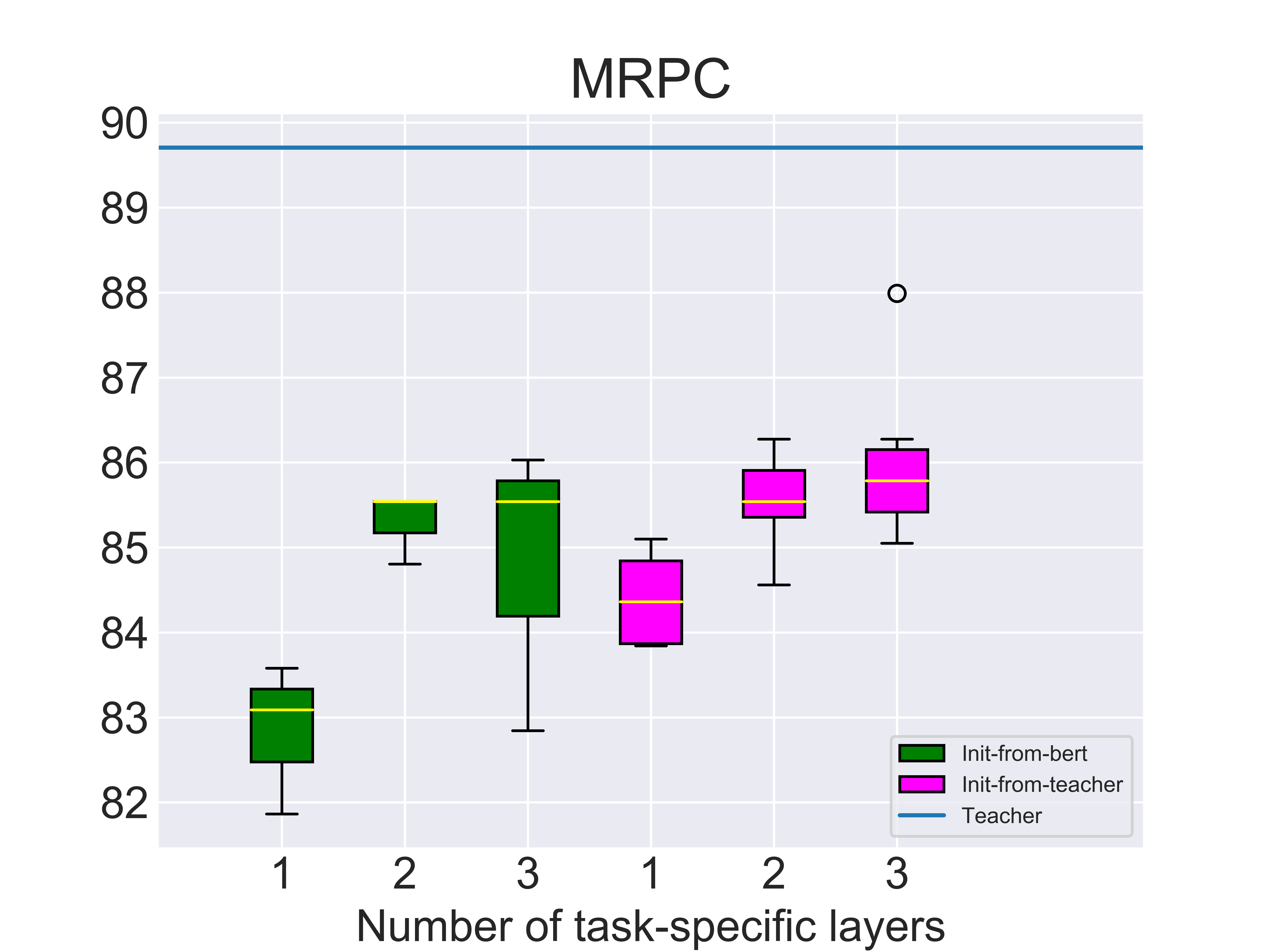}
     \end{subfigure}

\begin{subfigure}[b]{0.45\textwidth}
         \centering
         \includegraphics[width=\textwidth]{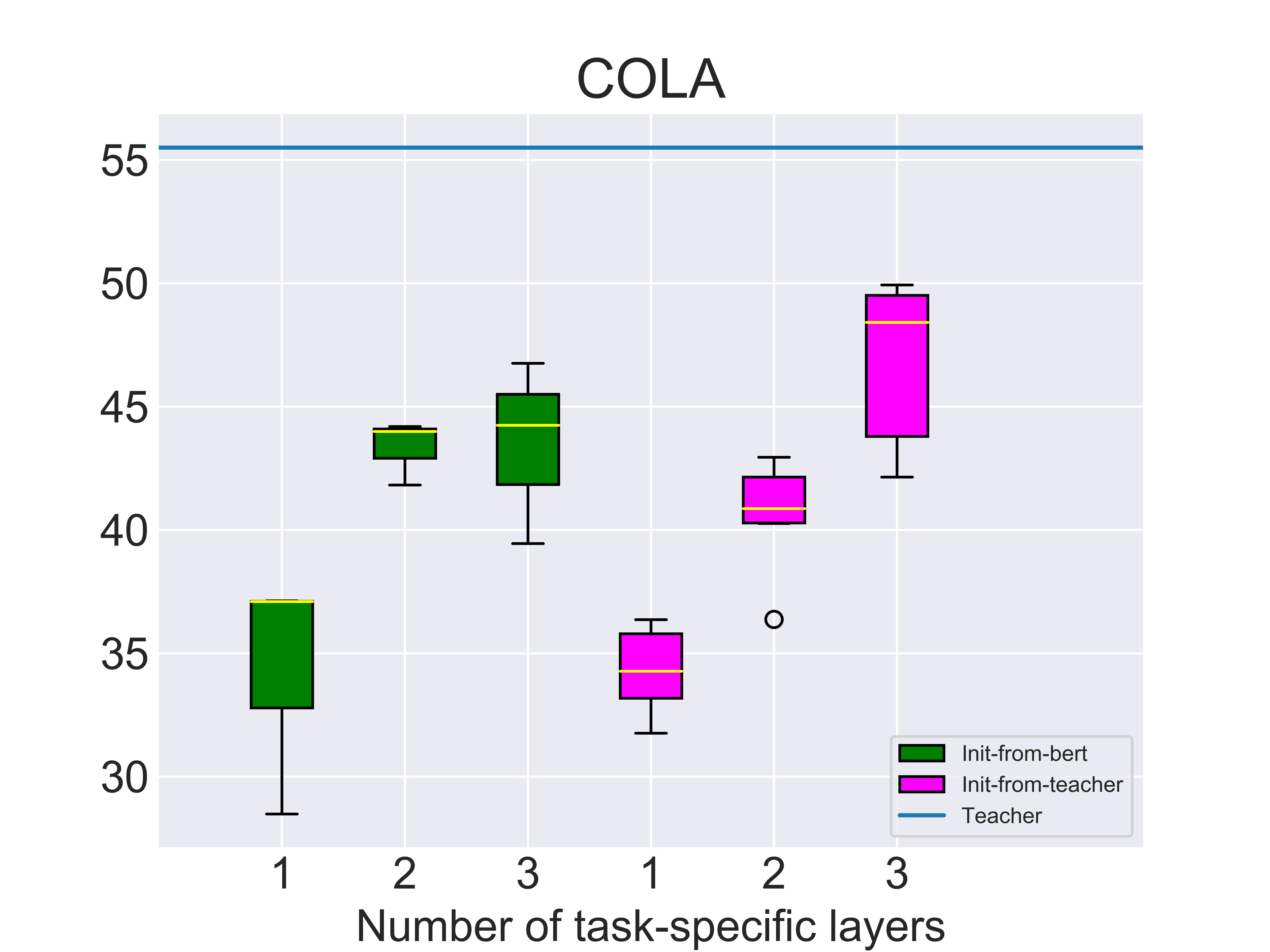}
     \end{subfigure}
     \begin{subfigure}[b]{0.45\textwidth}
         \centering
         \includegraphics[width=\textwidth]{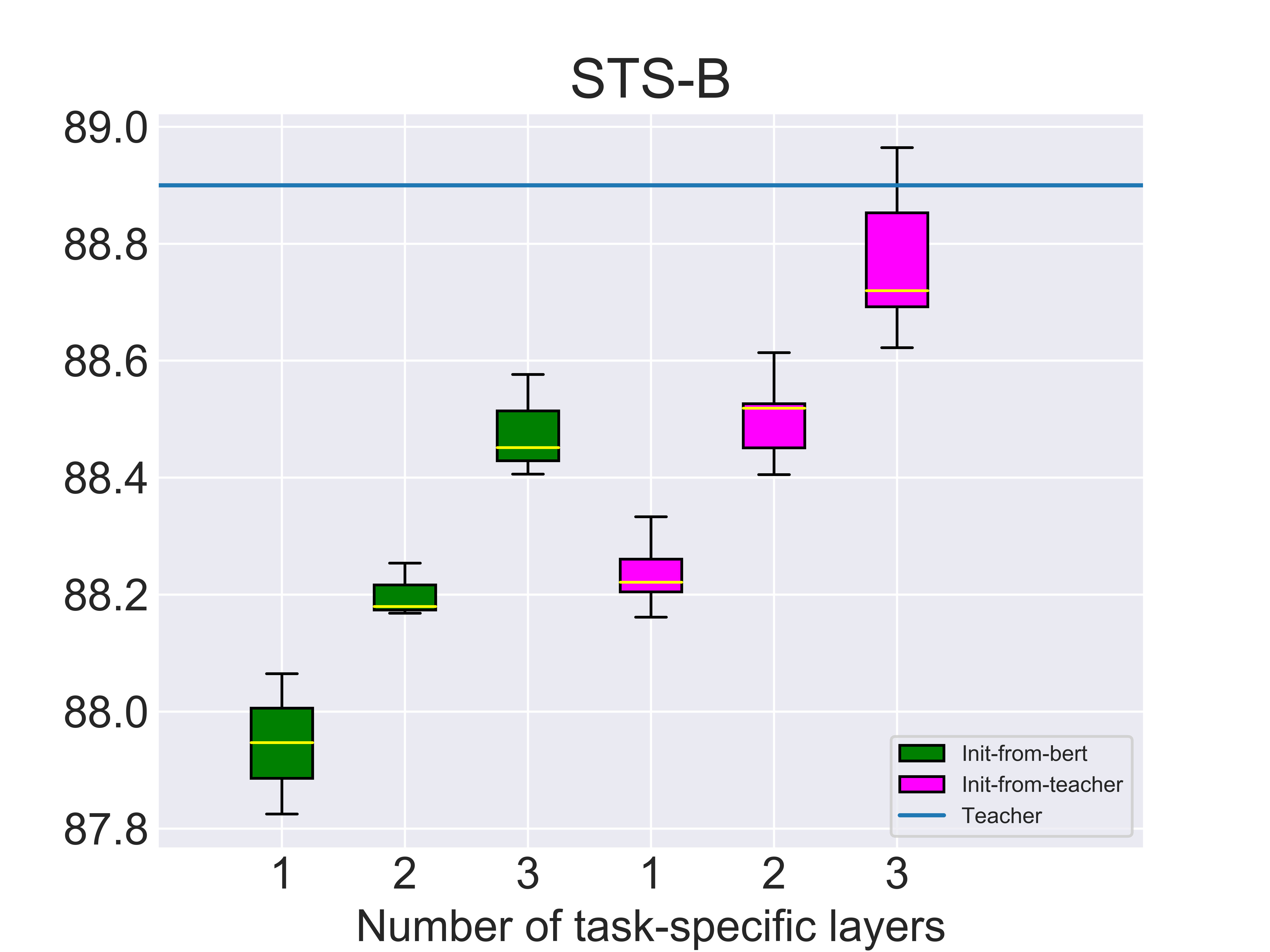}
     \end{subfigure}
\caption{A comparison of the task performance between vanilla initialization (initialize from pre-trained BERT) and teacher initialization as described in Section \ref{stkd} for $n\in\{1,2,3\}$, where $n$ is the number of task-specific layers in the student model. }
\label{ablation}
\end{figure*}

In the box plots of Figure \ref{ablation} above we report the performance of the student models initialized from pre-trained BERT and from the teacher.
It can be clearly seen that the latter initialization scheme generally outperforms the former. Besides, we also observe that although increasing the number of task-specific layers improves the performance, the marginal benefit of doing so varies across tasks. Notably, for QQP and STS-B the student models with only one task-specific layer are able to attain 99\% of the performance of their teacher.    

\section{Performance-Efficiency Trade-off \label{trade_off}}
In Fig \ref{tradeoff}, we report the performance of our method with various values of $c$, 
where $c$ is defined as the minimal marginal benefit (in terms of task performance metric) that every task-specific layer should bring
(see Section \ref{baselines}).

\begin{table*}[ht!]
\begin{center}
\resizebox{2.\columnwidth}{!}{%
\begin{tabular}{l|cccccccc|c||c|c}
\toprule
                         & QNLI         & RTE          & QQP         & MNLI          & SST-2       & MRPC         & CoLA      & STS-B    & Avg. & Layers       & Overhead \\
\midrule 
Full fine-tuning          &  \cb{91.6}  & \cb{69.7}    & \cb{91.1}   & \cb{\bf 84.6} & \cb{93.4}   & \cb{88.2}    & \cb{54.7} & \cb{88.8}   & \cb{82.8} & $12\times8$  & 96 (100\%) \\
\midrule 
Ours (KD-1)           & 86.4            & 66.1         & \cb{91.0}   & {77.5}        & {90.7}        & 85.1       & {36.4}    & {88.3}   & 77.4 & $7+1\times$8 & 15 (15.6\%) \\
Ours (KD-2)           & 88.6            & 64.6         & \cb{\bf 91.3} & {81.7}      & \cb{92.7}     & 86.3       & {44.0}    & \cb{88.6}   & 79.7 & $7+2\times$8 & 23 (24.0\%)  \\
Ours (KD-3)           & 90.2            & 66.8         & \cb{91.2}   & {82.9}        & \cb{92.7}     & \cb{88.0}  & {50.0}    & \cb{\bf88.9}   & 81.3 & $7+3\times$8 & 31 (32.3\%) \\
\midrule
Ours ($c=1.0$)          & 90.2            & \cb{71.5}    & \cb{91.0}   & {82.9}        & \cb{92.7}     & \cb{88.0}    & \cb{\bf 55.2} & \cb{88.3}   & 82.5 & $7+26$ & 33 (34.3\%) \\
                      & ${\scriptstyle(2, 3)}$       & ${\scriptstyle(7, 5)}$     & ${\scriptstyle(2, 1)}$   & ${\scriptstyle(4, 3)}$   
                      & ${\scriptstyle(6, 2)}$    & ${\scriptstyle(7, 3)}$     & ${\scriptstyle(4, 8)}$    & ${\scriptstyle(4, 1)}$      &  &  &  \\
Ours ($c=2.0$)          & 88.6            & {66.1}    & \cb{91.0}   & {81.7}        & \cb{92.7}     & {85.1}    & {50.0} & \cb{88.3}   & 80.4 & $7+13$ & 20 (20.2\%) 
\\
                      & ${\scriptstyle(2, 2)}$       & ${\scriptstyle(7, 1)}$     & ${\scriptstyle(2, 1)}$   & ${\scriptstyle(4, 2)}$   
                      & ${\scriptstyle(6, 2)}$    & ${\scriptstyle(7, 1)}$     & ${\scriptstyle(4, 3)}$    & ${\scriptstyle(4, 1)}$      &  &  &  \\
Ours ($c=3.0$)          & 86.4            & {66.1}    & \cb{91.0}   & {81.7}        & {90.7}     & {85.1}    & {50.0} & \cb{88.3}   & 79.9 & $7+11$ & 18 (18.8\%) 
\\
                      & ${\scriptstyle(2, 1)}$       & ${\scriptstyle(7, 1)}$     & ${\scriptstyle(2, 1)}$   & ${\scriptstyle(4, 2)}$   
                      & ${\scriptstyle(6, 1)}$    & ${\scriptstyle(7, 1)}$     & ${\scriptstyle(4, 3)}$    & ${\scriptstyle(4, 1)}$      &  &  &  \\
\midrule
Ours (w/o KD)         & {\bf \cb{91.7}}  & \cb{71.5}    & \cb{91.1}  & \cb{84.5}     & \cb{93.1}     & \cb{\bf 89.7} & \cb{\bf 55.2} & \cb{\bf 88.9}   & \cb{\bf 83.2} & $7+60$ & 67 (69.8\%) \\
                      & ${\scriptstyle(2, 10)}$       & ${\scriptstyle(7, 5)}$     & ${\scriptstyle(2, 10)}$   & ${\scriptstyle(4, 8)}$   
                      & ${\scriptstyle(6, 6)}$    & ${\scriptstyle(7, 5)}$     & ${\scriptstyle(4, 8)}$    & ${\scriptstyle(4, 8)}$      &  &  &  \\
\bottomrule
\end{tabular}
}
\end{center}
\caption{Results with various values of $c$. This parameter controls the performance-efficiency trade-off of the overall multi-task model, in the sense that 
we allow the growth of an existing task module by one more task-specific layer
only if that would bring a performance gain greater than $c$. \label{tradeoff}}
\end{table*}

\section{Industrial Application \label{industrial_application}}
We have implemented our framework in the application of utterance understanding of XiaoAI, a mono-lingual (Chinese) commercial AI assistant developed by XiaoMi. Our flexible multi-task model forms the bulk of the utterance understanding system, which processes over 100 million user queries per day with a peak throughput of nearly 4000 queries-per-second (QPS).

For each user query, the utterance understanding system performs various tasks, including emotion recognition, incoherence detection, domain classification, intent classification, named entity recognition, slot filling, etc.  Due to the large workload, these tasks are developed and maintained by a number of different teams.
As the AI assistant itself is under iterative/incremental development, its utterance understanding system undergoes frequent updates\footnote{Not necessarily frequent for any particular task, but overall frequent if we regard the system as a whole.}:
\begin{itemize}
\item Update of training corpus, e.g. when new training samples become available or some mislabeled samples are corrected or removed.
\item Redefinition of existing tasks. For instance, when a more fine-grained intent classification is needed, we may need to redefine existing intent labels or introduce new labels.
\item Introduction of new tasks. This may happen when the AI assistant needs to upgrade its skillsets so as to perform new tasks (e.g. recognize new set of instructions, play verbal games with kids, etc). 
\item Removal of obsolete tasks. Sometimes a task is superseded by another task, or simply deprecated due to commercial considerations. Those tasks need to be removed from the system. 
\end{itemize}
 One imperative feature for the system is the \emph{modular design}, i.e. the tasks should be independent of each other so that any modification made to one task
 does not affect the other tasks.
Clearly, a conventional multi-task system does not meet our need as multi-task training breaks modularity.

Before the introduction of BERT, our utterance understanding system is based on {single-task serving}, i.e. a separate model is deployed for each task. 
As those models are relatively lightweight (e.g. TextCNN, LSTM), overhead is not an issue.
However, with the introduction of BERT, the cost for single-task serving becomes a valid concern as each task model (a unique 12-layer fine-tuned BERT) requires two Nvidia Tesla V100 GPUs for stable serving that meets the latency requirement.

With the primary objective of reducing cost, we have implemented the proposed flexible multi-task model in our utterance understanding system, which provides serving for a total of 21 downstream tasks. 
Overall, there are 40 transformer layers of which 8 are shared frozen layers (on average $1.5$ task-specific layers per task). Using only 5 Nvidia Tesla V100 GPUs, we are able to achieve\footnote{with fp16 and fast transformer (\url{https://github.com/NVIDIA/FasterTransformer}) acceleration.} a P99 latency of 32 ms under a peak throughput of 4000 QPS. 
Compared with single-task serving for 21 tasks which would require 42 GPUs, 
we estimate that our system reduces the total serving cost by up to 88\%.
\newpage
\section*{Responsible NLP Research Checklist}
\subsection*{A. For every submission}

{A1. Did you discuss the \textit{limitations} of your work?}\\
\emph{Yes, it is explicitly discussed in Section \ref{limitations}.}\\

\noindent{A2. Did you discuss any potential risks of your work?}\\
\emph{No, we believe that there is no potential risk.}\\

\noindent{A3. Do the abstract and introduction summarize the paper's main claims?}\\
\emph{Yes, we confirm so.}\\

\subsection*{B. Did you use or create scientific artifacts?}
\emph{Yes, we used the GLUE datasets in Section \ref{experiments}.}\\

\noindent{B1. Did you cite the creators of artifacts you used?} \\
\emph{Yes, the GLUE paper is cited in Section \ref{experiments}. The individual datasets in GLUE are cited in Appendix \ref{glue_datasets}.}\\

\noindent{B2. Did you discuss the license or terms for use and/or distribution of any artifacts?}\\
\emph{No. Since those artifacts are popular in the NLP community, we merely followed the common practice of using these artifacts. We do not believe that our usage violate the license for use, or is potentially risky in any ways we can imagine. }\\

\noindent{B3. Did you discuss if your use of existing artifact(s) was consistent with their intended use, provided that it was specified? For the artifacts you create, do you specify intended use and whether that is compatible with the original access conditions (in particular, derivatives of data accessed for research purposes should not be used outside of research contexts)?}\\
\emph{No. The justification is the same that for question B2.}\\

\noindent{B4. Did you discuss the steps taken to check whether the data that was collected/used con- tains any information that names or uniquely identifies individual people or offensive content, and the steps taken to protect / anonymize it?
}\\
\emph{No. The justification is the same that for question B2.}\\

\noindent{B5. Did you provide documentation of the artifacts, e.g., coverage of domains, languages, and linguistic phenomena, demographic groups represented, etc.?
}\\
\emph{Yes, it is provided in Appendix \ref{glue_datasets}.}\\

\noindent{B6. Did you report relevant statistics like the number of examples, details of train/test/dev splits, etc. for the data that you used/created?
}\\
\emph{Yes, it is provided in Appendix \ref{glue_datasets}.}\\

\subsection*{C. Did you run computational experiments?}
\emph{Yes, in Section \ref{experiments}.} \\

\noindent{C1. Did you report the number of parameters in the models used, the total computational budget (e.g., GPU hours), and computing infrastructure used?
}\\
\emph{No, we did not report the number of parameters in the models used as it can be easily inferred from Table \ref{comparison}.  The total computation budget was  discussed in Section \ref{limitations}.}\\

\noindent{C2. Did you discuss the experimental setup, including hyperparameter search and best-found hyperparameter values?}\\
\emph{Yes, it is provided in the Section \ref{experiments} and Appendix \ref{hyper_parameters}.}\\

\noindent{C3. Did you report descriptive statistics about your results (e.g., error bars around results, summary statistics from sets of experiments), and is it transparent whether you are reporting the max, mean, etc. or just a single run?}\\
\emph{Yes, we explicitly stated in the caption of Table \ref{partial_freezing} and Table \ref{comparison} that our results are the maximum over 5 independent runs. 
Detailed results are also reported in Appendix \ref{detailed_results}.}\\

\noindent{C4. If you used existing packages (e.g., for preprocessing, for normalization, or for evaluation), did you report the implementation, model, and parameter settings used (e.g., NLTK, Spacy, ROUGE, etc.)?
}\\
\emph{We did reuse the WordPiece implementation from BERT's repository \url{https://github.com/google-research/bert} for tokenization. We did not report this as we consider it as a trivial matter.}\\

\subsection*{D. Did you use human annotators (e.g., crowdworkers) or research with human subjects?}
\emph{No, we did not use any human annotators, nor did we research with human subjects.}
\end{document}